\documentclass[journal]{IEEEtran}

\ifCLASSINFOpdf
\else
\fi
\usepackage{graphicx,fancyhdr,multirow,array,url,bm,amsmath,CJK,indentfirst}
\usepackage{mathrsfs}
\usepackage{color}
\usepackage{cite}
\usepackage{hyperref}

\usepackage{fancyhdr}

\begin{document}
\title{Rethinking deinterlacing for early interlaced videos}
\author{Yang~Zhao,~
Wei~Jia,~
Ronggang~Wang~

\thanks{This work is supported by the National Natural Science Foundation of China (Nos.61972129, 62072013, 62076086), and Shenzhen Research Projects of JCYJ20180503182128089, 201806080921419290, RC JC20200714114435057, GXWD20201231165807007-20200806163656003. R. Wang is the corresponding author.}
%
\thanks{Y. Zhao and W. Jia are with School of Computer and Information, Hefei University of Technology, Hefei 230009, China (e-mail: yzhao@hfut.edu.cn; jiawei@hfut.edu.cn).

R. Wang is with the School of Electronic and Computer Engineering, Peking University Shenzhen Graduate School, 2199 Lishui Road, Shenzhen 518055, China (e-mail: rgwang@pkusz.edu.cn).

Y. Zhao and R. Wang are also with Peng Cheng National Laboratory, Shenzhen 518000, China.
%
}}

\maketitle

\begin{abstract}
In recent years, high-definition restoration of early videos have received much attention. Real-world interlaced videos usually contain various degradations mixed with interlacing artifacts, such as noises and compression artifacts. Unfortunately, traditional deinterlacing methods only focus on the inverse process of interlacing scanning, and cannot remove these complex and complicated artifacts. Hence, this paper proposes an image deinterlacing network (DIN), which is specifically designed for joint removal of interlacing mixed with other artifacts. The DIN is composed of two stages, \emph{i.e.}, a cooperative vertical interpolation stage for splitting and fully using the information of adjacent fields, and a field-merging stage to perceive movements and suppress ghost artifacts. Experimental results demonstrate the effectiveness of the proposed DIN on both synthetic and real-world test sets.
\end{abstract}


%
\IEEEpeerreviewmaketitle

\section{Introduction}

Interlacing artifacts are commonly observed in many early videos, which are caused by interlacing scanning in early television systems, \emph{e.g.}, NTSC, PAL, and SECAM. As shown in Fig. 1, the odd lines and even lines of an interlaced frame are scanned from two different half-frames, \emph{i.e.}, the odd/top/first field and the even/bottom/second field. The two fields cannot be exactly aligned; hence, comb teeth aliasing appears in these different areas, especially when there are large movements between the two fields. With the rapid development of image restoration techniques, high-definition reconstruction of early videos has achieved impressive results. However, there are few studies about the interlacing artifacts removal that often appear in early videos.

Traditional deinterlacing methods \cite{1,2} are mainly designed for simple interlacing scanning process. Unfortunately, interlaced effects are often mixed with many other unnatural artifacts in real-world early videos, such as blocking and noise during compression, transfer, and storage. The performances of traditional deinterlacing methods thus often significantly decrease. Additionally, recent state-of-the-art (SOTA) image restoration models are not specifically designed for this artificial interlaced mechanism. As a result, the motivation of this paper is to specifically design an effective deinterlacing network for the joint restoration tasks of interlaced frames.

\begin{figure}[!t]
\centering
\graphicspath{{image/}}
\includegraphics[width=2.2 in]{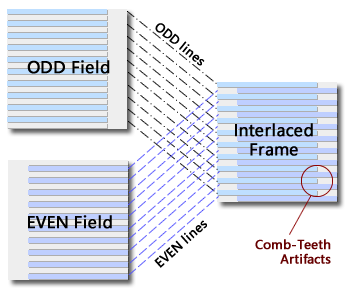}
\caption{Illustration of the interlaced scanning mechanism.}
\label{fig_1}
\end{figure}

\begin{figure*}[t]
\centering
\graphicspath{{image/}}
\includegraphics[width=6.0in]{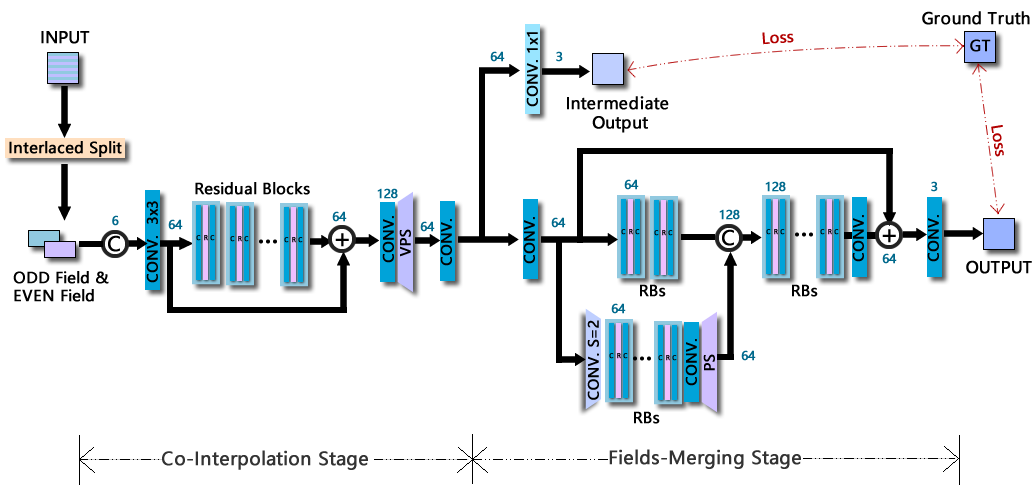}
\caption{Architecture of the proposed deinterlacing network.}
\label{fig_2}
\end{figure*}

The traditional interlaced scanning mechanism can be defined as $\bm{Y}=S(\bm{X_1},\bm{X_2})$, where $\bm{Y}$ denotes the interlaced frame, $S(\cdot)$ is the interlaced scanning function, and $\bm{X_1},\bm{X_2}$ denote the odd and even fields. Traditional deinterlacing methods focus on the reversed process of $S(\cdot)$, which can be roughly divided into four categories,\emph{ i.e.}, temporal interpolation, spatial interpolation, motion-based approaches, and learning-based methods. Temporal interpolation-based deinterlacing utilizes adjacent frames to interpolate the missing lines. However, the differences between two frames lead to visible artifacts when there are large movements in adjacent frames. Spatial interpolation is another kind of basic deinterlacing technique that adopts only one field to estimate the whole frame. The simplest spatial interpolation is the line average and edge line average. Various single field interpolation methods have been proposed in the past decades, for instance, edge detection with fuzzy logic \cite{3}, block-wise autoregression \cite{4}, bilateral filtering model \cite{5}, locality and similarity adaption \cite{6}, the maximum a posteriori model \cite{7} and least square methods \cite{8}. To further refine the visual quality, visual saliency is also introduced to improve deinterlacing, for example, spectral residual visual saliency-based deinterlacing \cite{9,10}. There are also some single field interpolation methods for special applications, such as GPU-parallel implementation of edge-directed interpolation \cite{11} and deinterlacing for 1080p-to-8KUHD \cite{12}. To improve deinterlacing methods for large movements, motion-based methods have been proposed to make use of two fields \cite{13,14,15} or adaptively improve single field methods \cite{16,17}. Many motion-compensation-based methods have also been introduced to calculate motion vectors between two frames to estimate the movements. Proper temporal or spatial interpolations are then applied according to the estimated movements \cite{18,19,20,37,38}. Recent dictionary learning-based models have also been applied for deinterlacing, \emph{e.g.}, Choi et al. \cite{21} used a joint dictionary learning method to reconstruct interlaced high dynamic range video. With the explosion of deep neural network (DNN)-based methods in many low-level vision tasks \cite{22,23,27,28,29}, Zhu et al. \cite{30} first introduced a lightweight deinterlacing convolutional neural network (DICNN) to deinterlacing task based on a shallow super-resolution CNN (SRCNN) architecture \cite{23}. Liu et al. \cite{35} further improved DICNN by using a deformable convolution and attention-residual-blocks. Bernasconi et al. \cite{36} presented an effective multi-field deinterlacing method based on residual dense network. In \cite{34}, Akyüz proposed a residual network with dilated convolutions for joint deinterlacing and denoising of dual-ISO HDR images. However, these approaches are mainly focused on reverse of interlacing scanning and still cannot handle the complex and complicated artifacts in real-world early videos. In this paper, the degradation process of interlaced early video is computed as $\bm{Y}=C(S(\bm{X_1},\bm{X_2}))+\bm{n}$, where $C(\cdot)$ denotes the video compression and $\bm{n}$ represents multiple noise. Hence, the proposed image deinterlacing method should jointly consider the complex interlacing artifacts mixed with other compression artifacts and noise.

To address this problem, a deinterlacing network (DIN) is specifically designed based on the analysis of the traditional interlacing scanning mechanism. The proposed DIN consists of two stages: the co-interpolation stage and the field-merging stage. In the co-interpolation stage, the input frame is first divided into odd field and even field, and then the two fields are simultaneously inputted into a ResNet to implement vertical interpolation. Hence, the network can clearly know that these two fields are from two different frames and can still make use of the mutual latent features since these two frames have many similar contents. Additionally, the vertical pixel-shuffling module is adopted to simulate the interlaced scanning process. In the field-merging stage, the interpolated field features are fused to the final output. The ResNet structure is selected to avoid the loss of high-frequency details. Note that larger receptive fields are needed in this stage for perceiving the movements and removing the ghost artifacts of the merged frame. Another downscale and reconstruction branch is thus added to achieve a larger receptive field. Note that it is difficult to obtain ground truth for early videos, and we train the proposed DIN on a synthetic datasets and test it on both the synthetic datasets and real-world early video frames.

\section{The Proposed Method}
\subsection{Architecture of the Deinterlacing Network}

Different from many image restoration tasks that have blind degradation models, interlaced scanning is a fixed artificial mechanism. Hence, the network is specially designed according to the characteristics of interlaced scanning. The architecture of the proposed DIN is illustrated in Fig. 2, which consists of two stages, \emph{i.e.}, the co-interpolation stage (CIS) and the fields-merging stage (FMS). In the following, each stage is introduced and analysed in detail.

\subsection{Co-Interpolation Stage}

The main purpose of CIS is to compute the vertical interpolation of two fields, and its structure is illustrated in detail in Fig. 2. An effective enhanced super-resolution network (EDSR) \cite{27,28} is adopted as the backbone of CIS, which utilizes a typical ResNet structure \cite{31} and carefully optimizes the residual blocks for image restoration tasks. Note that we do not use ResNet as a blackbox to entirely reconstruct interlaced frame, and do not first recover the interlaced fields and then apply interlacing scanning as in DICNN \cite{30} either. In the proposed DIN, the input frame is first split into even field and odd field according to the interlacing mechanism, which contain even lines and odd lines, respectively. There are mainly two benefits of this split and co-interpolation subnetwork. First, instead of the whole frame that contains serious comb teeth artifacts, the odd and even fields are used as input. Hence, the disruption of interlacing artifacts is minimized during the training of the DIN. Second, the two adjacent fields have considerable similar spatial information. As a result, the two fields are inputted simultaneously into the network, which is helpful to recover the high-frequency pixel-level details.

Additionally, the split fields have half the size of the original frame. Motivated by the efficient pixel-shuffling (PS) module \cite{26}, which is widely adopted in current SOTA super-resolution methods \cite{24,27,28} to reduce the computational cost, the CIS also reconstructs the fields in low-resolution space and then utilizes a vertical pixel shuffling (VPS) at the end of the network to magnify the reconstructed features. Similar to the traditional interlacing scanning mechanism in Fig. 1, the VPS module interlacedly combine odd lines and even lines into an entire frame, which can be regarded as a special case of a normal PS module.

In our experiment, the CIS contains $6$ residual blocks (RB). Each RB consists of a $3\times3$ convolutional layer (conv.), a rectified linear unit (ReLU) activation, another conv. $3\times3$, and a local residual connection, as in \cite{27}. The common global skip is used to improve the convergence of deep ResNet.

\subsection{Fields-Merging Stage}

The most significant artifact in the deinterlacing process is caused by the large movement between two fields. Hence, the subnetwork of the FMS tends to have a large receptive field to perceive movement information. Note that some commonly used architectures with large receptive fields, \emph{e.g.}, UNet and autoencoder, often lead to the loss of pixelwise spatial details. Hence, the proposed FMS applies multiscale ResNet, which utilizes a base ResNet to maintain pixel-level information and then adds a downscale branch to perceive larger-scale movements. The final feature of the downscale branch is magnified via a PS module and then concatenated to the middle of the base ResNet.

By means of the two-scale structure, the proposed FMS can effectively preserve high-frequency details and extract differences at a larger scale simultaneously. As shown in Fig. 2, in our experiment, 3 RBs are used in the downscale branch; conv. $3\times3$ with stride $2$ is utilized as the downsampling module, and the $2\times2$ PS module is adopted for upsampling after RBs. The base branch contains $5$ RBs, which have $2$ RBs before the concatenation of the downscale branch, and $3$ RBs after the concatenation. The global skip connection is also used. In general, adding more scales can extract higher-level semantic information and perceive larger movement. However, the parameters and computational cost also increase accordingly. In our experiment, we have observed that two-scale FMS can already handle the deinterlacing task well. Hence, for keeping simplicity, only two scales are used in the proposed DIN.

\subsection{Loss Function of the DIN}

Similar to many image restoration models, the proposed DIN is end-to-end trained via a $L_1$-distance-based loss function. As shown in Fig. 2, the two stages of the DIN are both supervised by the ground truth data. To constrain the interpolated field features by means of single ground truth, a temporary conv. $1\times1$ is used to linearly combine the interpolated field features (64 channels) to the intermediate output. This supervision of intermediate output is added to directly optimize the CIS module and better train the co-interpolation stage. The entire network is then optimized by means of the following loss function:
\begin{equation}
\begin{split}
L=\frac{1}{N_T}\sum \limits_{n=1}^{N_T}[\lambda||C_1(F_{CIS}(\bm{X_1^n},\bm{X_2^n}))-\bm{Y^n}||_1+\\
(1-\lambda)||F_{FMS}(F_{CIS}(\bm{X_1^n},\bm{X_2^n}))-\bm{Y^n}||_1],
\end{split}
\end{equation}
where $N_T$ denotes the total number of training samples, $\bm{X_1^n},\bm{X_2^n}$ are the odd and even fields, $\bm{Y^n}$ represents the ground truth frame, $C_1$ is the conv. $1\times1$ operation, $F_{CIS}$ and $F_{FMS}$ denote subnetworks of CIS and FMS, respectively, and $\lambda$ is an artificial weight of the intermediate supervision. This weight $\lambda$ is initially set as 0.5, so the CIS module can be constrained to recover better vertical interpolation results. Then, the $\lambda$ is gradually reduced to 0.1 in the last half of the training phase.

\subsection{Implementational Details}

In our experiment, a total of $100,000$ training patch pairs with a size of $256\times256$ are selected from the training set, and then these patches are augmented via horizontal and vertical flipping. It should be noted that rotation cannot be used for data augmentation in deinterlacing because the interlaced artifacts are along the horizontal direction. In this paper, each conv. layer in RBs outputs $64$ feature maps except for the last $3$ RBs in FMS, which have $128$ channels after the concat operation. The weights are initialized via Xavier, and the network is optimized via Adam. The initial learning rate is $10^{-4}$ and then decreases by a factor of $10$ after every $20$ epochs. This network was trained $100$ epochs with a GTX1080Ti GPU by means of the MXNET platform.

\begin{figure*}[ht]
\centering
\graphicspath{{image/}}
\includegraphics[width=6.0in]{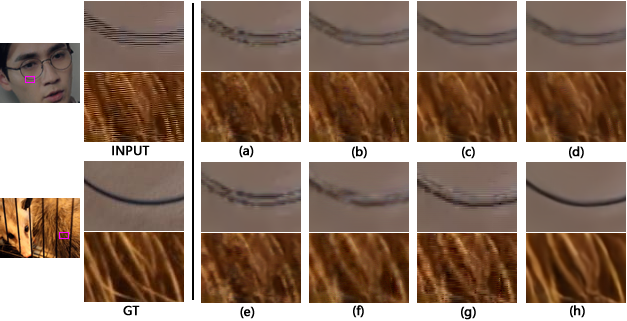}
\caption{Deinterlacing results on synthetic YOUKU-2K dataset, (a) retrained SRCNN \cite{23} (b) retrained EDSR \cite{27}, (c) retrained RCAN \cite{24}, (d) retrained RDN \cite{25}, (e) DICNN \cite{30}, (f) retrained DJDD \cite{34}, (g) retrained STCLN \cite{35}, (h) the proposed DIN.}
\label{fig_3}
\end{figure*}

\begin{figure*}[ht]
\centering
\graphicspath{{image/}}
\includegraphics[width=6.4in]{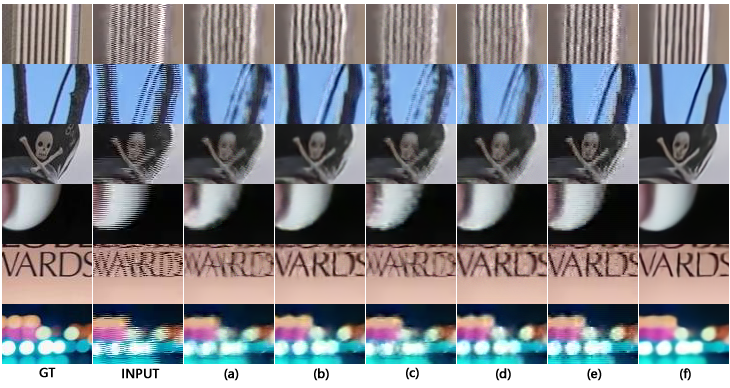}
\caption{Deinterlacing results on synthetic REDS and VIMEO90K datasets, (a) retrained EDSR \cite{27}, (b) retrained RDN \cite{25}, (c) DICNN \cite{30}, (d) retrained DJDD \cite{34}, (e) retrained STCLN \cite{35}, (f) the proposed DIN.}
\label{fig_4}
\end{figure*}

\section{Experiments}
\subsection{Training and Testing Datasets}

For training, a total of 400 video sequences are collected from the YOUKU-2K video dataset \cite{32}. Each video sequence consists of 100 frames with 2K resolution. Then, the interlaced frame is produced according to the traditional interlaced scanning system, \emph{i.e.}, the odd and even lines are scanned from two adjacent frames. The real first frame is utilized as ground truth data. Furthermore, the interlaced video is then compressed via the H.264 codec with a random \emph{``-crf''} parameter from 30 to 38 by means of FFMPEG tool. This compression process is adopted to roughly simulate the mixed compression artifacts in early videos.

For testing, we used three synthetic test sets and a real-world early video dataset. In the synthetic test sets, 40 frames from YOUKU-2K dataset, 75 frames from REDS validation set \cite{40} and 225 frames form VIMEO90K dataset \cite{39} are randomly selected for testing. All the testing frames are generated in the same way as in the training stage. In the real-world dataset, 38 interlaced frames are randomly selected from several real-world early videos to verify the generalization of the DIN.

\begin{figure*}[ht]
\centering
\graphicspath{{image/}}
\includegraphics[width=6.8in]{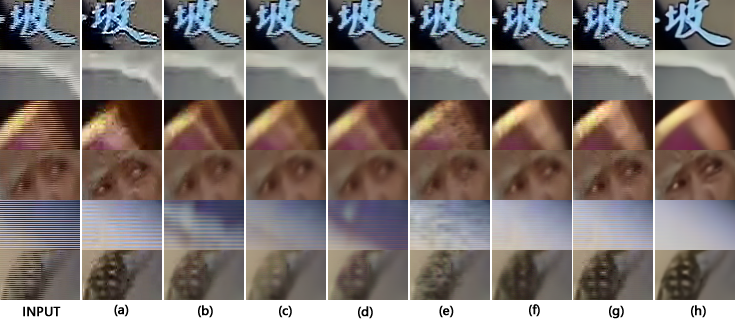}
\caption{Reconstruction results of real-world early video frames with different methods, (a) single field EDSR \cite{27}, (b) retrained EDSR \cite{27}, (c) retrained RCAN \cite{24}, (d) retrained RDN \cite{25}, (e) DICNN \cite{30}, (f) retrained DJDD \cite{34}, (g) retrained STCLN \cite{35}, (h) the proposed DIN.}
\label{fig_5}
\end{figure*}

\subsection{Experimental Results on the Synthetic Testing Sets}
In our experiment, the proposed DIN is compared with deep deinterlacing models of DICNN \cite{30}, DJDD (deep joint deinterlacing and denoising network) \cite{34} , STCLN (spatial-temporal correlation learning network) \cite{35}, and several image restoration networks, e.g., super-resolution models SRCNN \cite{23}, EDSR \cite{27}, RCAN \cite{24}, and denoise model RDN \cite{25}. For a fair comparison, these networks are all retrained with the same training set. Note that the PS upsampling module in EDSR and RCAN is not used because the resolution is not magnified. In addition, single field interpolation is also compared by using EDSR \cite{27} as upsampling method. 

Fig. 3 illustrates the results of several interlaced and compressed frames in YOUKU-2K set. First, retrained SOTA models of EDSR, RCAN, RDN and DJDD can reduce interlacing and compression noise, but they still cause blur or ghost artifacts. Second, lightweight DICNN and STCLN, which are designed for deinterlacing without noise, directly combine reconstructed fields with original noisy fields. Therefore, they cannot remove these mixed artifacts. Last, the DIN can remove interlacing and compression artifacts simultaneously.

\begin{table}[!htp]
\centering
\caption{PSNR and SSIM results on the synthetic YOUKU-2K dataset}
\scriptsize
\begin{tabular}{m{2.2cm}m{1.0cm}m{1.0cm}*{1}{m{2.2cm}<{\centering}}}
\hline
~ &\textbf{PSNR} &\textbf{SSIM}  &\textbf{Number of Param.} \\
\hline
\hline
Input	              &30.33	&0.8777  &-\\
\hline
Single Field EDSR	  &32.04	&0.9088  &5.905M\\
\hline
SRCNN (retrain)	\cite{23}    &32.43	     &0.9111   &\textbf{0.036M}\\
\hline
EDSR (retrain) \cite{27}	     &32.65	     &0.9130  &5.905M\\
\hline
RCAN (retrain) \cite{24}              &32.45	&0.9168 &5.066M\\
\hline
RDN (retain) \cite{25}              &32.60	&0.9178 &2.375M\\
\hline
DICNN \cite{30}              &32.53	&0.9125  &\underline{0.118M}\\
\hline
DJDD \cite{34}	&\underline{33.23}	&\underline{0.9211}	&0.925M\\
\hline
STCLN \cite{35}	&32.84	&0.9124	&0.351M\\
\hline
DIN	                &\textbf{35.17}	  &\textbf{0.9348}  &1.816M\\
\hline
\hline
\end{tabular}
\end{table}

\begin{table}[!htp]
\centering
\caption{PSNR and SSIM results on the synthetic REDS and VIMEO90K datasets}
\scriptsize
\begin{tabular}{m{2.0cm}*{4}{m{1.0cm}<{\centering}}}
\hline
~ &\multicolumn{2}{c}{\textbf{REDS}} &\multicolumn{2}{c}{\textbf{VIMEO90K}} \\
\cline{2-5}
~ &\textbf{PSNR} &\textbf{SSIM}  &\textbf{PSNR} &\textbf{SSIM} \\
\hline
\hline
Input	              &24.79	&0.7059	&25.05	&0.7697\\
\hline
EDSR(retrain) \cite{27}	&26.14	&0.7437	&26.94	&0.8291\\
\hline
RCAN(retrain) \cite{24}	&27.49	&0.7797	&29.66	&0.8690\\
\hline
RDN(retrain) \cite{25}	&\underline{27.81}	&\underline{0.7828}	&\underline{29.87}	&\underline{0.8714}\\
\hline
DICNN \cite{30}	         &26.06	&0.7397	&26.64	&0.8208\\
\hline
DJDD \cite{34}	          &26.56 &0.7549	&28.91	&0.8548\\
\hline
STCLN \cite{35}	         &27.27	&0.7573	&29.25	&0.8390\\
\hline
DIN	                &\textbf{28.17}	&\textbf{0.8007}	&\textbf{30.34}	&\textbf{0.8827}\\
\hline
\hline
\end{tabular}
\end{table}

Fig .4 shows results on REDS and VEMIO90K test sets. The RDN and DJDD can reproduce better results for large motion than EDSR and shallow DICNN. However, they still cannot suppress complex artifacts mixed with interlacing. The STCLN can reduce ghost artifact by means of deformable convolution, but it still reproduces severe noises. By comparing the visual quality, the proposed DIN obtains cleaner and more natural results than these SOTA models.

Related objective assessments are listed in Table \uppercase\expandafter{\romannumeral 1} and Table \uppercase\expandafter{\romannumeral 2}. The DIN outperforms other DNN-based methods on these synthetic test sets. Note that the DICNN and STCLN are designed for real-time deinterlacing task, which are very lightweight base on shallow backbones. By comparing with directly applying deeper model of EDSR, RCAN and RDN, which are not specifically designed for joint deinterlacing task, the DIN performs more effective with fields split, co-interpolation and fields merging mechanism. 

\subsection{Experimental Results on Real-World Early Videos}
Fig. 5 shows the reconstructed results of real-world early video frames. There is considerable aliasing and compression noise in these frames. First, the single field EDSR avoids comb teeth but also magnifies aliasing during vertical interpolation. Second, the lightweight DICNN and STCLN cannot totally remove mixed noise and aliasing. Third, the SOTA models of EDSR, RCAN and RDN can reduce interlacing and compression artifacts, however, they still lead to ghost shadows and blurry results. Fourth, the DJDD can suppress ghost artifacts, but the results of DJDD also contain blur or noise. Finally, although these models are trained on the same dataset, the DIN can significantly remove interlacing mixed with noises, which demonstrates that the proposed method is effective for these real-world early videos.

Single stimulus subjective testing has been applied to compare the visual quality of reconstructed early videos. The details of the viewing conditions and test session are set according to Rec. ITU-R BT.500 \cite{33}. A total of 15 observers are invited to score the impairment scales of six sets of frames to 5 levels, where the impairment levels 5, 4, 3, 2, and 1 denote ``imperceptible impairment'', ``perceptible impairment'', ``slightly annoying impairment'', ``annoying impairment'', and ``very annoying impairment'', respectively. The average impairment scale scores (ISS) are listed in Table \uppercase\expandafter{\romannumeral 3}. It can be easily found that the proposed method achieves much higher subjective scores than other methods.
Scores of two DNN-based blind image quality assessment (BIQA) methods, \emph{i.e.}, HyperIQA \cite{41} and KonIQ \cite{42}, are also listed in Table \uppercase\expandafter{\romannumeral 3}. Compared with other SOTA models, DIN can achieve higher results. Note that these BIQA methods are not trained with interlaced degradations. Therefore, these BIQA scores are only used for rough comparisons, and are not completely consistent with visual quality.
In addition, average running time of different methods is also listed in Table \uppercase\expandafter{\romannumeral 3}. Lightweight DICNN and STCLN run much faster than other deep models. By comparing with other deeper models, the DIN can achieve better results with less time cost. (More results can be found in supplementary materials: https://gitee.com/gityzhao/din/)

\begin{table}[!t]
\centering
\caption{BIQA and Impairment scale scores on real-world early videos}
\scriptsize
\begin{tabular}{m{2.2cm}*{2}{m{1.2cm}<{\centering}}*{2}{m{1.0cm}<{\centering}}}
\hline
~ &\textbf{HyperIQA} &\textbf{KonIQ} &\textbf{ISS} &\textbf{Time (s)} \\
\hline
\hline
Input	&28.47    &41.68  &2.14	  &-\\
\hline
EDSR (retrain) \cite{27}	&29.08 &41.34  &2.85	&0.77\\
\hline
RCAN (retrain) \cite{24}	&28.91 &41.67 &3.12	&1.34\\
\hline
RDN (retrain) \cite{25}	&28.81 & 41.63 &2.82	&1.44\\
\hline
DICNN \cite{30}	&\underline{29.09} &41.69 &2.54	&\textbf{0.08}\\
\hline
DJDD \cite{34}	&28.95 &41.67 &\underline{3.65}	&1.16\\
\hline
STCLN \cite{35}	&29.04  &\underline{41.82} &2.70	&\underline{0.12}\\
\hline
DIN	 &\textbf{29.32} &\textbf{42.91} &\textbf{4.33}	&0.73\\
\hline
\hline
\end{tabular}
\end{table}

\section{Conclusion}
This paper proposes a deinterlacing network (DIN) for early videos that contains interlacing and mixed compression artifacts. The two-stage architecture of the proposed DIN is motivated by the interlacing scanning mechanism and traditional deinterlacing strategies. The co-interpolation stage splits the interlaced frame into odd and even fields and then estimates the vertical interpolation of two fields. Another field-merging stage is then applied to further remove ghost shadows and other unnatural artifacts. Experimental results demonstrate that the DIN can effectively recover high-quality results from interlaced and compressed frames.


\ifCLASSOPTIONcaptionsoff
\newpage
\fi

\end{document}